\documentclass{article}
\usepackage[top=1in, bottom=1in, left=1in, right=1in]{geometry}
\usepackage[utf8]{inputenc}

\usepackage{hyperref}
\usepackage{url}
\usepackage{overpic}
\usepackage{booktabs}
\usepackage{tabularx}
\usepackage{thm-restate}
\usepackage{amsmath}
\usepackage{amsfonts}
\usepackage{float}

\usepackage{xcolor}

\newcommand{\ay}[1]{\textcolor{black}{#1}}
\newcommand{\ayy}[1]{\textcolor{black}{#1}}

\title{T-SHRED: Symbolic Regression for Regularization and Model Discovery with Transformer Shallow Recurrent Decoders}

\author{Alexey Yermakov$^{1,2,*}$, David Zoro$^{1,*}$, Mars Liyao Gao$^{3}$, and J. Nathan Kutz$^{^{1,2}}$\\[.1in]
{ $^1$ Electrical and Computer Engineering, University of Washington, Seattle, WA }\\
{ $^2$ Applied Mathematics, University of Washington, Seattle, WA  }\\
{$^3$ Computer Science \& Engineering, University of Washington, Seattle, WA }\\
{$^{*}$Co-first authors}\\[.2in]
\texttt{\{alexeyy,zorodav,marsgao,kutz\}@uw.edu} \\
}

\date{\today}

\begin{document}

\maketitle

\begin{abstract}
SHallow REcurrent Decoders (SHRED) are effective for system identification and forecasting from sparse sensor measurements. Such models are light-weight and computationally efficient, allowing them to be trained on consumer laptops. SHRED-based models rely on Recurrent Neural Networks (RNNs) and a simple Multi-Layer Perceptron (MLP) for the temporal encoding and spatial decoding respectively. Despite the relatively simple structure of SHRED, they are able to predict chaotic dynamical systems on different physical, spatial, and temporal scales directly from a sparse set of sensor measurements. In this work, we modify SHRED by leveraging transformers (T-SHRED) embedded with symbolic regression for the temporal encoding, circumventing auto-regressive long-term forecasting for physical data. This is achieved through a new sparse identification of nonlinear dynamics (SINDy) attention mechanism into T-SHRED to impose sparsity regularization on the latent space, which also allows for immediate symbolic interpretation. Symbolic regression improves model interpretability by learning and regularizing the dynamics of the latent space during training. We analyze the performance of T-SHRED on three different dynamical systems ranging from low-data to high-data regimes. 
%We observe that the SINDy-Attention mechanism outperforms standard Multi-Head Self-Attention in T-SHRED models in next-step forecasting across all tested datasets due to its interpretable symbolic model.
\end{abstract}

\section{Introduction}

Advancements in science and engineering have always relied on the discovery of symbolic expressions, which aid in the fundamental understanding of a system.  Starting from the computational work on the BACON software package in the late 1970s by Langley~\cite{langley1977bacon}, symbolic regression algorithms aim to use a diversity of regression techniques to search through a space of mathematical expressions to find the model that best fits a given dataset, both in terms of accuracy and simplicity.  Symbolic regression can be extended to looking for relationships among the derivatives of variables, which is equivalent to finding governing differential or partial differential equations (e.g. space and time derivatives).  Currently, there are a number of symbolic regression software packages, which aim to assist scientific discovery by finding relationships among variables in a data set or governing equations of motion~\cite{schmidt2009distilling,cranmer2023interpretable,brunton2016discovering,udrescu2020ai,jin2019bayesian,landajuela2022unified,kim2020integration,la2021contemporary,muthyala2025symantic,petersen2019deep}.  Here, we demonstrate the power of symbolic regression for the purpose of interpretability and regularization of the latent space of deep learning algorithms, most notably transformers.  Thus, symbolic regression, specifically the sparse identification of nonlinear dynamics (SINDy)~\cite{brunton2016discovering}, is used to improve the interpretability of the ubiquitous transformer network by imposing dynamical systems models on the attention heads of shallow recurrent decoders.

The SHallow REcurrent Decoder (SHRED) architecture has been demonstrated to be an effective deep learning model for scientific and engineering applications. SHRED is based on three key concepts: (i) the separation of variables technique for solving partial differential equations~\cite{tomasetto2025reducedordermodelingshallow}, (ii) Taken's embedding theorem~\cite{takens}, and (iii) a decoder (spatial) only architecture.  The model can be seen as a joint training to learn the temporal trajectory and spatial field of the input data simultaneously through the temporal encoder and spatial decoder units respectively~\cite{williams2024sensing,tomasetto2025reducedordermodelingshallow}. Prior works have used SHRED to perform state space reconstruction from a sparse set of sensors in the spatial dimension with application in broad scientific domains~\cite{williams2024sensing, tomasetto2025reducedordermodelingshallow, gao2025sparseidentificationnonlineardynamics,kutz2024shallow,introini2025models,ebers2024leveraging,mei2024long,ni2024wavefield}. 
These works have shown that a full state reconstruction of the data can be obtained from a randomly placed set of sparse sensors where the sensor count can be as little as 3 to reconstruct the original space of over $\mathcal{O}(10^7)$ spatial measurements. 
Furthermore, the SHRED models are agnostic to the specific system they are modeling. They can perform Go-Pro physics, where dynamics are learned directly from video \cite{gao2025sparseidentificationnonlineardynamics, tomasetto2025reducedordermodelingshallow}; they can also learn chaotic fluid dynamics from simulations \cite{tomasetto2025reducedordermodelingshallow, gao2025sparseidentificationnonlineardynamics}.

While it is clear that the theoretical motivation for SHRED models has led to promising results, its architecture does not take advantage of recent deep learning advancement for temporal data which include transformer models.
First used for machine translation \cite{vaswani2017attention}, the self-attention mechanism in the transformer is able to learn complex patterns from large datasets \ay{by acting on the input as a fully connected graph of tokens \cite{joshi2025transformers}}. Since \ay{their inception}, significant empirical evidence has been produced showing that transformer-based models scale exceptionally well with more data in a rich variety of domains and architectures \cite{Zhai_2022_CVPR, kaplan2020scaling, liang2024scaling}. \ay{Transformers are powerful encoder functions that have been used extensively in foundation models due to their capacity to capture complex interactions between input elements while simultaneously being hardware efficient \cite{DBLP:journals/corr/abs-2108-07258}}.

The attention mechanism, on the other hand, is not optimally designed for physics modeling with symbolic understanding. 
Recall that the attention mechanism is based on mapping a query with a key-value pair~\cite{vaswani2017attention}. 
A strong motivation for the attention mechanism is to query information from rich historically available data~\cite{wang2025learning,zhong2025understanding}. 
To improve the performance of transformers on physics and time-series learning, we leverage the observation that the attention mechanism can be interpreted as a numerical ODE solver for the convection-diffusion equation in a multiple-particle dynamical system~\cite{lu2019understanding,chen2018neural,agarwal2024iterated,sander2022sinkformers}. 
Building upon the dynamical systems perspective, linear attention~\cite{zeng2023transformers,katharopoulos2020transformers,dao2024transformers} provides a compelling alternative that not only achieves strong empirical performances, but also aligns well with the structure of physical laws.
This connection motivates us to further incorporate symbolic regression to capture a broader range of dynamics, with the potential to generalize across a wider range of physical systems. 

In this paper, we introduce Transformer-SHRED (T-SHRED) which uses a transformer backbone within the SHRED architecture for sparse sensor modeling. 
%With the power of the transformer, T-SHRED builds upon the original SHRED architecture to model latent space dynamics and outperforms the naive SHRED transformer model.
We also introduce SINDy-Attention, which embeds a symbolic regression unit into each attention head, outperforming the traditional self-attention layer \ay{and unlocking stable one-shot long-term forecasting}. 
SINDy-Attention enforces the learning of structured, generalizable dynamics in the latent space, incorporating complex physics directly into the model. 
With SINDy-Attention, T-SHRED not only learns a much better latent dynamics, but also produces scientific discovery via the interpretable latent dynamical system.  \ay{Specifically, it learns a parsimonious relationship between the time derivative and the learned variables, much as one would construct when deriving governing equations.  This has been one of the dominant forms of scientific discovery throughout history~\cite{kutz2022parsimony}.}

\begin{figure}[t]
\centering
\includegraphics[width=0.9\textwidth]{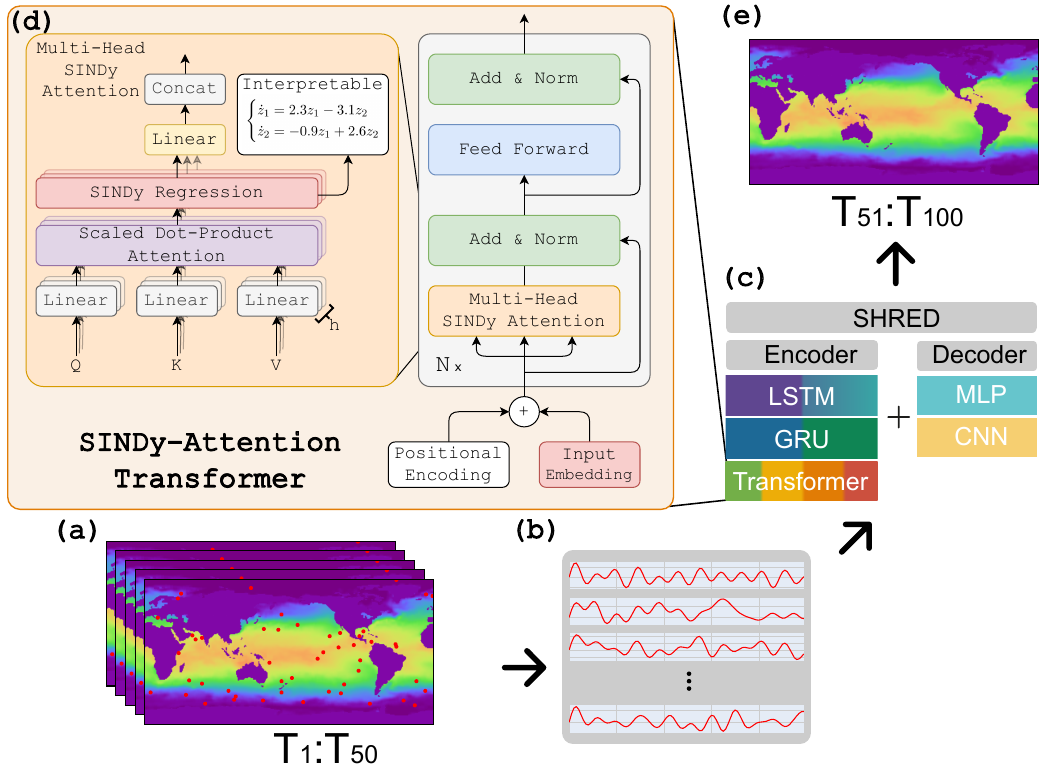}
\caption{Illustration of the SHRED architecture. The SHRED architecture takes a time-series of sparse sensor measurements (a,b) from spatio-temporal data and outputs the next-step full state prediction (e). The SHRED architecture consists of an encoder and a decoder (c), optionally with SINDy Loss during training and SINDy attention for the transformer encoder. The T-SHRED architecture is a SHRED model with a transformer encoder and either an MLP or CNN decoder. The T-SHRED encoder is the transformer architecture modified to include a SINDy Regression layer in the self-attention heads (d). This forces the model to learn an interpretable model of the dynamics of the latent space. Each head is capable of learning a separate ODE in the latent space (d).}\label{fig:t_shred}
\end{figure}

We perform a comparative study using three datasets to evaluate the performance of T-SHRED in \ay{short-term and long-term} full-state prediction from a sparse set of measurements. 
These datasets come from various complex physical phenomena on different scales and with varying dataset sizes. 
We show that T-SHRED with SINDy-Attention performs well on the next-step state prediction task and on long-term forecasting \ay{while avoiding auto-regressive rollouts}.
We also demonstrate that SINDy-Attention T-SHRED enables model interpretability with symbolic expressions.
For a complete set of experiments, we also study the effect SINDy loss \cite{gao2025sparseidentificationnonlineardynamics} has on our prediction task as well as using a Convolutional Neural Network (CNN) as the decoder for each model.
In summary, we conduct comprehensive experiments to demonstrate the effectiveness and interpretability of T-SHRED across diverse datasets and model configurations.

The contribution of this paper is three-fold: 
\begin{itemize}
    \item We propose T-SHRED, a SHRED model with a transformer unit with various available decoders (MLP and CNN decoders). 
    T-SHRED is able to learn complex dynamics and allows the shallow decoder to reconstruct fine details \ay{and provides stable long-term rollouts}.
    \item With symbolic regression as a regularization, the SINDy-Attention mechanism regularizes the attention heads of the transformer to learn interpretable dynamics in the latent space. 
    \item A comprehensive set of experiments is conducted on complex dynamical datasets to compare the effectiveness of T-SHRED with previous SHRED models, and we provide the discovered dynamical system embedded within the SINDy-Attention head. 
\end{itemize}

This paper is organized as follows. In \autoref{sec:background} we review the history of SHRED models. We motivate the proposed T-SHRED and the symbolic SINDy-Attention mechanism in \autoref{sec:tshred}. Our proposed approach is then bench-marked on challenging datasets in \autoref{sec:experiments}. In \autoref{sec:conclusion} we conclude and discuss future directions. Our code is publicly available at \href{https://github.com/yyexela/T-SHRED}{https://github.com/yyexela/T-SHRED}.

\section{Background} \label{sec:background}

\subsection{Shallow recurrent decoder network (SHRED)}
SHRED is a spatial learning architecture that estimates high-dimensional states from limited temporal sensor measurements~\cite{williams2024sensing}. 
The first SHRED networks pass a sequence of sensor measurements into an LSTM to create a latent space representation of the temporal dynamics which are then mapped to the full state space through a Shallow Decoder Network \cite{williams2024sensingshallowrecurrentdecoder}. The SHRED architecture is based on the separation of variables technique, which assumes that the solution to a PDE can be decomposed into the product of spatial and temporal functions ${u(x,t)=T(t)X(x)}$~\cite{tomasetto2025reducedordermodelingshallow}. Then, solving the PDE simplifies to solving two separate ODEs: ${dT/dt=cT}$ and ${\mathcal L X = cX}$ where $\mathcal  L$ is the linear operator for the spatial derivatives, modeling the underlying physics of the system and $c$ is some constant. The insight with SHRED is that while these two differential equations are separable, they are not independent since they are related through the constant $c$. In SHRED, the learning of $T(t)$, $X(x)$, and $c$ is carried out through the joint training of a neural network architecture of the form ${u=X(T(\{y_i\}_{i=t-k+1}^{t}))}$, where $t$ are the time steps in the training data and $k>0$ is the temporal lag of the model for measurements $y_i$. For large enough $k$, there is theoretical evidence that SHRED is able to learn the full spatial field at a given point in time from sparse sensor measurements~\cite{williams2024sensing}. This result extends to coupled PDEs where multi-dimensional fields are capable of being reconstructed.  
SHRED can also be thought of as a generalization of dynamic mode decomposition which can be used to produce uncertainty quantification estimates by statistical bagging~\cite{sashidhar2022bagging} or potentially multiscale models~\cite{brunton2019methods}.  

\subsection{SINDy-SHRED}

% \ay{AY: Mars, what is $^{(i)}$ doing for $\Xi$? Is it necessary? Please define variables $h$, $i$, and $k$.}
% Thanks I deleted {(i)} and h, k are defined I found..? 

SINDy-SHRED~\cite{gao2025sparseidentificationnonlineardynamics} improves on the core SHRED architecture by regularizing the latent space variables with the Sparse Identification of Nonlinear Dynamical Systems (SINDy) algorithm~\cite{brunton2016discovering}, similar to what can be done with encoding and decoding schemes~\cite{gin2021deep,mars2024bayesian,champion2019data,lusch2018deep,otto2019linearly}. 
Symbolic regression as a latent space regularization is done by treating the encoder's latent space variables as a system of differential equations composed of a library of polynomial and trigonometric terms. These coefficients are then added to the loss function of the model during training. The SINDy library coefficients $\Xi$ are optimized in an ensemble during training by minimizing the following:

\begin{align}
\Xi &= \arg \min \left| \left| \mathbf z_{t+1} - \left ( \mathbf z_{t} + \sum_{j=0}^{k-1} \Theta (\mathbf z_{t+jh}) \Xi h \right ) \right| \right|_2^2+ ||\Xi||_0,
\end{align}
where $\mathbf{z}_{t+jh} = \mathbf z_t + \Theta (\mathbf z _{t+(j-1)}h) \Xi h $ are the intermediate steps for forward simulation, $\{\mathbf{z_i}\}_{i=1}^T$ is the trajectory data in the latent space, and $h = \frac{\Delta t}{k}$ defines the time-step for forward Euler integration each $\Delta t$ with $k$ mini-steps. Notice that $\ell_0$ regularization enforces sparsity in the library coefficients. In practice, however, SINDy-SHRED utilizes $\ell_2$ regularization with pruning, which is an approximation to $\ell_0$ under regularity conditions \cite{gao2023convergence,zheng2014high}.  
The trained SINDy-SHRED model produces an interpretable ODE model that describes the dynamics of the given physical phenomena.

\section{Transformer SHRED}\label{sec:tshred}

% \subsection{SL-TRED}
% \subsection{SA-TRED}
% \subsection{SASL-TRED}

\label{sec:method}

Informed by advances in deep learning, we modified the SHRED architecture by replacing the recurrent encoder with a transformer. We also study the impact of replacing the shallow decoder with a convolutional decoder.

\subsection{Transformers} Transformers are a powerful and efficient alternative to sequence modeling~\cite{vaswani2017attention,elman1990finding,hochreiter1997long,schuster1997bidirectional,beck2024xlstm}, they leverage parallel computations and utilize a self-attention mechanism to capture long-range dependencies in time-series data. The attention mechanism enables the transformer to identify the most relevant parts of the inputs for the downstream task which is informed by the loss function. The core of the transformer encoder is the Multi-Head Self-Attention (MHSA) mechanism, which computes attention as follows. Given an input $\mathbf{x} \in \mathbb{R}^{n \times d}$, for each head $h$:
\begin{align}
    Q^{(h)}(\mathbf{x}) &= \mathbf{x} W_{h,q}, \quad K^{(h)}(\mathbf{x}) = \mathbf{x} W_{h,k}, \quad V^{(h)}(\mathbf{x}) = \mathbf{x} W_{h,v}, \\
    \text{Attention}^{(h)}(\mathbf{x}) &= \text{softmax} \left( \frac{Q^{(h)}(\mathbf{x}) {K^{(h)}(\mathbf{x})}^T}{\sqrt{k}} \right) V^{(h)}(\mathbf{x}),
\end{align}
where $W_{h,q}, W_{h,k}, W_{h,v} \in \mathbb{R}^{d \times k}$ are learnable parameters. The outputs from all heads are concatenated and passed through a feed-forward network.

Multiple attention heads allow the model to learn multiple latent representations of the input. The weight matrix $W_o\in\mathbb{R}^{Hd\times d}$ is a learnable set of weights that allow the model to learn how to combine the various heads:
\begin{align}
    \text{MHSA}(\mathbf{x}) &= \text{Concat}(\text{Attention}^{(1)}(\mathbf{x}), \dots, \text{Attention}^{(H)}(\mathbf{x})) W_o, \\
    \tilde{\mathbf{x}} &= \text{LayerNorm}(\mathbf{x} + \text{MHSA}(\mathbf{x})),
\end{align}
where the layer normalization is applied to stabilize the training with a skip-connection. 
Finally, there is another feed-forward network with another skip connection and layer normalization:
\begin{align}
    \text{MLP}(\mathbf{\tilde{\mathbf{x}}}) &= \sigma(\tilde{\mathbf{x}} W_1) W_2, \\
    \mathbf{z} &= \text{LayerNorm}(\tilde{\mathbf{x}} + \text{MLP}(\tilde{\mathbf{x}})),
\end{align}
where $W_1, W_2^T \in \mathbb R^{d \times m}$ are learnable parameters of the feed-forward network, and $\sigma(\cdot)$ is the Rectified Linear Unit activation function. The final output $\mathbf{z}$ represents the transformed input after passing through one transformer layer.

Transformers have become the state-of-the-art in many machine learning domains~\cite{islam2023comprehensivesurveyapplicationstransformers}, including language modeling, biomedical imaging and spatio-temporal modeling, due to their scalability and ability to handle large datasets. 
In T-SHRED, we apply a transformer to perform temporal prediction instead of an LSTM network. 
This improves generalization through multiple attention heads and enables the model to learn complex temporal relationships.

\subsection{Convolutional Neural Networks (CNNs)}

CNNs have been widely applied in computer vision tasks which are designed to capture local patterns and spatial hierarchies in images via convolutions.
In T-SHRED, we replace the shallow decoder with a CNN decoder, which can effectively model enhanced and fine-grained spatial features in the output state space.

The formulation of the CNN decoder is as follows. Given the latent space representation $\mathbf{z} \in \mathbb{R}^{n \times d}$, the CNN decoder applies a series of convolutional layers to produce the output state space $\mathbf{y} \in \mathbb{R}^{n \times m}$:
\begin{align}
    \mathbf{y} &= \text{Conv}(\mathbf{z}) = \sigma{(\text{Conv}_1(\sigma(\text{Conv}_2(...\sigma(\text{Conv}_{\ell}(\mathbf{z}))...))))},
\end{align} 
where $\text{Conv}_\ell$ represents the $\ell$-th convolutional layer, $m$ is the desired output dimension and $\sigma$ is the Rectified Linear Unit activation function. The CNN decoder captures complex spatial features in the output state space, enhancing the model's ability to reconstruct high-dimensional states from limited sensor measurements.

\subsection{Latent Space Symbolic Regression and Physics Regularization}

Different attention mechanisms reveal different characteristics of the transformer architecture~\cite{dao2022flashattention,katharopoulos2020transformers,sun2023retentive,dao2024transformers,guo2025log}. The attention mechanism in transformers can be interpreted as a dynamical systems modeling approach, where each attention head learns a different aspect of the temporal dynamics~\cite{geshkovski2023mathematical,lu2019understanding,phuong2022formal}.
To see this, we reformulate the attention mechanism as the following. We first notice that the attention mapping is an $\mathbb{R}^d\to\mathbb{R}^d$ function. This is explicitly characterized as the following system~\cite{geshkovski2023mathematical}:
\begin{equation}
\begin{split}
    \dot{\mathbf{x}}_i(t) = P_{\mathbf{x}_i(t)}^\perp \biggr ( \sum_{h=1}^H \sum_{j=1}^n &Z_{\beta,i,h}(t) e^{\beta \langle Q_h(t) \mathbf{x}_i(t), K_h(t) \mathbf{x}_j(t) \rangle} V_h(t) \mathbf{x}_j(t)+ w_t\sigma(a_t \mathbf{x}_i(t) + b_t)\biggr )
\end{split}
\end{equation}
where $P_{\mathbf{x}_i(t)}^\perp$ is the orthogonal projection onto the subspace orthogonal to $\mathbf{x}_i(t)$, $Z_{\beta,i,h}(t)$ is a normalization factor, and $w_t, a_t, b_t$ are learnable functions with input $t$.

In total we have $H$ attention heads, and each attention head learns different $Q_h(t)$, $K_h(t)$, and $V_h(t)$ weights that govern the dynamics of the latent space $\mathbf{x}_i(t)$. The attention mechanism can be viewed as a system of ordinary differential equations (ODEs) that evolve over time, where each head learns a different ODE governing the dynamics of the latent space. 
When having the continuous formulation above in a discrete setting using a Lie-Trotter splitting scheme, we obtain the standard procedure of the transformer~\cite{geshkovski2023mathematical}. 

This reformulation shows that the attention mechanism can be viewed as a system of ordinary differential equations (ODEs) that evolve over time, where each head learns a different ODE governing the dynamics of the latent space. This perspective aligns with the SINDy framework, which aims to discover governing equations from data, making it a natural fit for incorporating symbolic regression into the transformer architecture.

While the widespread adoption of black-box models in machine learning has led to significant improvements in predictive performance, they also create a lack of transparency and interpretability, making it
difficult for users to understand what the models are learning. In response, the literature has seen an increase in alternative approaches which focus on developing interpretable models by fitting data directly to understandable equations. One of the most notable of these approaches is SINDy \cite{brunton2016discovering, fasel2022ensemble}, which performs sparse symbolic regression on a library set of functions to find the governing equations of a dynamical system. SINDy has expanded into a subfield where a variety of techniques have been applied to solve domain specific problems \cite{zanna2020data, kaptanoglu2021physics, dam2017sparse, callaham2022empirical, guan2021sparse, loiseau2018constrained, loiseau2018sparse}. Motivated by these approaches, we adjust the self-attention mechanism in transformers to force each head of the self-attention mechanics to learn the dynamics of the latent space. 

We build upon the dynamical systems perspective of a transformer to introduce SINDy-Attention. SINDy-Attention regularizes each head of the transformer by fitting the latent space variables into a coupled ODE through sparse coefficient regularization. In particular, SINDy-Attention is a parameterized function of the form $f_\theta (\mathbf x) = \mathbf z : \mathbb R ^{n \times d} \rightarrow \mathbb R ^{n \times d}$ with learnable parameters $\theta$ for $\mathbf x \in \mathbb R^{n \times d}$. \ay{In each attention head, we utilize the differentiable ODE\_int($\cdot$, $\cdot$, $\cdot$) function from the \texttt{torchdiffeq} package to perform latent space rollouts, allowing for flexible forecasting \cite{torchdiffeq}}. Let $H \in \mathbb N$ be the number of heads in the transformer such that ${\exists k, H \cdot k = d}$ \ay{and $n_f$ be the number of time-steps to forecast the inputs.} Then, for $W_{h,q}, W_{h,k}, W_{h,v} \in \mathbb R^{d \times k}$, $\Xi^{(h)} \in \mathbb R^{\ell \times k}$, and $W_{ff_2}^T, W_{ff_1} \in \mathbb R^{d \times m}$, the SINDy-Attention transformer performs the following operation:

% x_i is R^d
% i: is n
% j: is dim

% Q,K,V(xj) is n x k
% S(h) is n x k
% S is n x d

% Theta: n x l
% Library: l x k

\begin{align}
    Q^{(h)}(\mathbf x) &= \mathbf x W_{h,q}, \quad K^{(h)}(\mathbf x) = \mathbf x W_{h,k}, \quad V^{(h)}(\mathbf x) = \mathbf x W_{h,v}\\
    {T^{(h)}} &= \text{rowsoftmax} \left( \frac{Q^{(h)}(\mathbf x) {K^{(h)}}(\mathbf x)^T}{\sqrt{k}} \right) V^{(h)}(\mathbf x) \\
    S^{(h)} &= \ay{\text{ODE\_int} \left(g: Y \rightarrow i Y \Xi^{(h)}, \Theta_{\text{SINDy}} \left({T^{(h)}}' \right), 1:n_{f}:1 \right)} \label{eq:MHSyA} \\
    S &= \text{concatenate}(S^{(1)}, \ldots, S^{(H)})\\
    \mathbf z &= (S\ W_{ff_1}) W_{ff_2}
\end{align}

\ay{The function $\Theta_{\text{SINDy}} : \mathbb R^{n \times k} \rightarrow \mathbb R^{n \times \ell}$ applies a library of $\ell$ functions to the latent space of some input matrix. Furthermore, ODE\_int($\cdot$, $\cdot$, $\cdot$) takes as input (1) the ODE, (2) the initial condition, and (3) the number of steps to forecast the ODE (here, we forecast from $1$ to $n_f$ with a step size of $1$). The learnable parameters $\theta$ are the weight matrices $W$ as well as the coefficients for the library $\Xi^{(h)}$. As seen in \autoref{eq:MHSyA}, the standard multi-head attention block is modified to do sparse regression on the latent space of each transformer head. We impose the constraint that the learned parameters in the latent space $\Xi ^{(h)}$
are symmetric and imaginary, resulting in purely imaginary eigenvalues and resulting in stable rollouts in the latent space.} After training, the individual heads of each T-SHRED layer can be interpreted as a coupled ODE that describes the dynamics of the latent space.

\subsubsection{Connection to SINDy-SHRED}
In SINDy-SHRED, the latent space variables are regularized with a library of polynomials and Fourier terms. 
By considering the attention mechanism as a dynamical system, we can see that a single head SINDy-Attention can be interpreted as a SINDy-unit applied within the learning structure. 
\ay{To illustrate the connection, consider the case of \textbf{single-head} attention, where the dynamical system underlying a transformer block can be compactly expressed as follows:}
\ay{
\begin{align}
    \tilde{\mathbf x}_{\text{transformer}} = \text{LayerNorm}(\text{MLP}(\mathbf x+\text{Attention}(\mathbf x))),
\end{align}
where the Attention unit can be either MHSA or SINDy-Attention here.
}
\ay{
SINDy-SHRED is a simplification of the above model by
\begin{align}
    \label{eqn:transformer_dynamical_system}
    \tilde{\mathbf x}_{\text{SINDy}} = \mathbf x+\Theta_{\text{SINDy}}(\mathbf x)\Xi\Delta t,
\end{align}
where $\Theta_{\text{SINDy}}(\cdot)$ contains the library of functions; $\Xi$ denotes SINDy coefficient; and $\Delta t$ is the step size for forward integration. 
}

SINDy-Attention, similar to SINDy-SHRED, utilizes a SINDy library to model the governing physics, and uses an MLP to capture higher-order interactions which are potentially missed in the SINDy library. 
Furthermore, it applies layer normalization to keep the dynamical system within a stable region. 
Without the LayerNorm operation and MLP block, the dynamical system modeling strategy of SINDy-Attention mechanism can be re-written into a variant of SINDy-SHRED~\cite{gao2025sparseidentificationnonlineardynamics}. 
\ay{This connection shows that SINDy-Attention extends the SINDy-SHRED framework by (i) enabling the modeling of physical interactions missing from the SINDy library, (ii) providing stable numerical integration schemes, and (iii) handling complex phenomena governed by multiple distinct physical laws through multi-head attention. 
It also shows that SINDy-SHRED can be interpreted as a simplified transformer architecture, which helps to explain the exceptional performance in~\cite{gao2025sparseidentificationnonlineardynamics}.
}

\section{Computational Experiments}
\label{sec:experiments}

\begin{figure}[H]
\centering
\includegraphics[width=\textwidth]{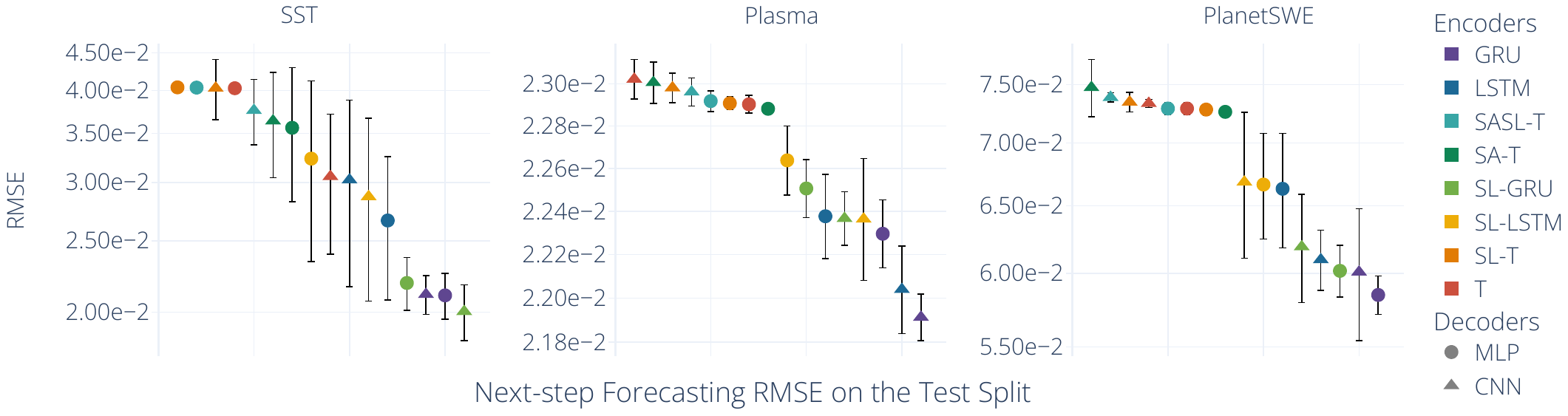}
\caption{Ranking of the best performing SHRED models on each dataset for nexts-step state prediction. The y-axis is the RMSE of next-step forecasting on the test dataset. SL-LSTM represents a SINDy-Loss LSTM, SL-GRU represents a SINDy-Loss GRU, T represents a Vanilla Transformer, SL-T represents a SINDy-Loss Transformer, SA-T represents the Sindy-attention Transformer, SASL-T a SINDy-Attention and SINDy-Loss Transformer.}\label{fig:scatter}
\end{figure}

This section explores three different physical systems that we train the SHRED models on to perform state prediction \ay{and forecasting}. These dynamical systems come from different physical scales such as particle physics and global weather phenomena, as well as low-data and high-data regimes.
\ay{To compare T-SHRED with the existing SHRED models, we perform a comprehensive model comparison by (1) evaluating the performance of every encoder and decoder on next-step state prediction from a sparse set of sensor measurements; and (2) comparing the performance of long-term rollouts on the lowest test set RMSE T-SHRED model with the lowest test set RMSE previous SHRED model. We perform this analysis for each of the three selected datasets.}

\ay{We preprocess each dataset by a linear scaling of each field to be in the range [0, 1] and splitting each track into a training, validation, and testing split. The training split takes the first 80\% of each track, the validation split takes the next 10\%, and the testing split takes the last 10\%.}

\ay{We compare each combination of 8 encoders and 2 decoders. The 8 encoders are a GRU, an LSTM, a vanilla transformer, and a SINDy-Attention transformer, with each encoder having a SINDy-Loss variant. The 2 decoders are an MLP and CNN.
The SINDy-Attention models are set to have a forecast length of 1 and use latent space polynomial order of 1 without a bias term, which is required to keep $\Xi^{(h)}$ symmetric. \ayy{Explicitly, the library $\Theta_{\text{SINDy}}$ for each SINDy-Attention head is linear, which necessarily results in linear ODEs in the latent space for all tested datasets.} All transformer models have 4 heads in their attention mechanism. All models use a hidden dimension of 20.
For the rest of the hyperparameters, we perform extensive hyperparameter tuning. For each model, we randomly select 5 sensor locations in the state space that persist across time as input to the SHRED models. We then train each encoder-decoder combination 25 times using the Multiobjective Tree-structured Parzen Estimator (MOTPE) algorithm for hyperparameter optimization with RayTune, Optuna, and the ASHA scheduler \cite{liaw2018tune, optuna_2019, li2020massivelyparallelhyperparametertuning}. Each training run uses a batch size of 128 and 50 epochs. The hyperparameter search space is described in \autoref{tab:hyperparams}, where we tune over the parameter dropout rate, the encoder depth, the learning rate, the SINDy-Loss weight, and the SINDy-Attention weight. For models that do not use SINDy-Loss or SINDy-Attention, we do not tune those loss weights. Note that for SINDy-Attention Transformers, only the final layer is set to have SINDy-Attention. The rest of the transformer encoder layers have the standard Multi-Head Self-Attention. All transformer-based models use an MLP to create input embeddings for the sensor inputs to the hidden dimension.}

\ay{The loss function for each model is shown in \autoref{eqn:loss}. Each model's loss function contains a forecast Mean Squared Error (MSE) term (\autoref{eqn:loss_forecast}), where $\mathbf{X}_t$ is the input sequence of sensors, $\hat  {\mathbf{Y}}_{t+1}$ is next-step full state space, $f_e$ and $f_d$ are the SHRED encoder and decoder respectively. For models with SINDy-Loss, a SINDy-Loss term is added (\autoref{eqn:loss_sl}). Similarly, for models with SINDy-Attention, a SINDy-Attention term is added, which regularizes the sum of the SINDy-Attention coefficients (\autoref{eqn:loss_sa}).}

\ay{For each training run in hyperparameter optimization, the full loss function is used on the training split. However, the hyperparameter optimization algorithm searches for the lowest validation loss equal to \autoref{eqn:loss} with $\alpha=0$ and $\beta=0$, corresponding to a pure reconstruction loss metric on the validation split. This setup allows each model to regularize its own architecture but standardizes the comparison to strictly be based on next-step state prediction. Once the hyperparameter optimization is completed, a final test loss is obtained for each encoder-decoder combination on the testing split (with $\alpha=\beta=0$). We then take the square root of the test loss to obtain a test split root-mean-squared error (RMSE). Note that the test loss is computed over the normalized data. We perform the complete hyperparameter optimization loop with 10 different seeds and report the mean and standard deviation of the final testing RMSE. To ensure a fair comparison, the sensor positions are fixed across seeds.
}

\ay{To forecast previous SHRED models, we auto-regressively feed the outputs of the model back in as an input. To forecast SINDy-Attention T-SHRED, we set $n_f$ from \autoref{eq:MHSyA} to the desired forecast length and do a single pass through the model from the inputs.}

\begin{figure}[H]
\centering
\includegraphics[width=\textwidth]{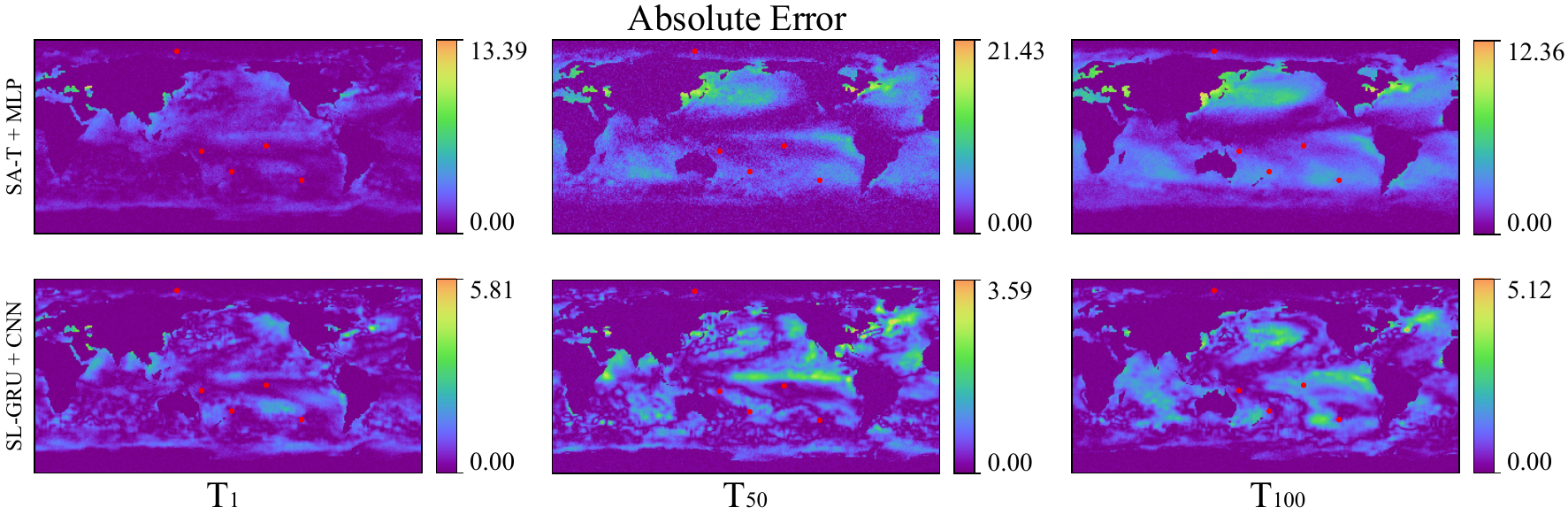}
\caption{Absolute error between the predicted and true values for the best performing SHRED and T-SHRED models on the Sea-Surface Temperature dataset. The top row shows the absolute error of T-SHRED with a SINDy-Attention encoder and an MLP decoder. The bottom row shows the absolute error of SHRED with a SINDy-Loss GRU encoder and a CNN decoder. The columns represent the timesteps in the testing set for which the models were evaluated. Sensor locations are in red.}\label{fig:sst_forecast}
\end{figure}

\begin{subequations}\label{eqn:loss}
\begin{align}
    \mathcal{L} &= \mathcal{L}_{\text{forecast}} + \alpha \cdot \mathcal{L}_{\text{SINDy--L}} + \beta \cdot\mathcal{L}_{\text{SINDy--A}}  \\
    \mathcal{L}_{\text{forecast}} = &\left|\left|f_d(f_e(\mathbf X_t)) - \hat {\mathbf Y}_{t+1}\right|\right|_2^2 \label{eqn:loss_forecast} \\
    \mathcal{L}_{\text{SINDy--L}} = &  \left| \left| \mathbf z_{t+1} - \left ( \mathbf z_{t} + \sum_{i=0}^{k-1} \Theta (\mathbf z_{t+ih}) \Xi h \right ) \right| \right|_2^2+ ||\Xi||_0 \label{eqn:loss_sl}\\
    \mathcal{L}_{\text{SINDy--A}} = &  \sum_{t=1}^{T} \sum_{h=1}^{H} \left|\left| \Xi_{(t)}^{(h)} \right|\right|_2^2 \label{eqn:loss_sa}\\
\end{align}
\end{subequations}

\subsection{Sea Surface Temperature}

The first dataset we consider is the Sea-Surface Temperature dataset (SST) dataset: a collection of 1,400 weekly snapshots of the weekly mean sea surface temperature collected from 1992 to 2019 by NOAA \cite{reynolds2002improved}. The data comes in a 180 $\times$ 360 grid with 44,219 of the 68,400 locations containing sea surface temperature information. The dataset's total size is 179MB.

\ay{The lowest average RMSE on the test split of $2.00 \cdot 10^{-2}$ ($\pm\ 1.74 \cdot 10^{-3}$) was achieved from SHRED with a SINDy-Loss GRU encoder and a CNN decoder with an average model size of 15.11 MB. The best performing T-SHRED model was a SINDy-Attention transformer encoder and an MLP decoder with an average RMSE on the test split of $3.56 \cdot 10^{-2}$ ($\pm\ 7.35 \cdot 10^{-3}$) and an average model size of 5.20 MB. A comparison of $100$ steps of forecasting between SHRED and T-SHRED is presented in \autoref{fig:sst_forecast}.}

\label{sec:hyperparams}
\begin{table}[h]
\begin{center}
\begin{tabular}{lccc}
\hline
Hyperparameter & Type & Min & Max \\
\hline
dropout & uniform & 0.0 & 0.1 \\
encoder\_depth & rand\_int & 1 & 10 \\
learning rate & log\_uniform & 0.0001 & 0.1 \\
sindy\_attention\_weight & log\_uniform & 1.0 & 10.0 \\
sindy\_loss\_weight & log\_uniform & 1.0 & 10.0 \\
\hline
\end{tabular}
\caption{Hyperparameter search space for all SHRED models. If a model is not a T-SHRED model, sindy\_attention\_weight is not tuned. Similarly for sindy\_loss\_weight if a model does not have SINDy-Loss.}
\label{tab:hyperparams}
\end{center}
\end{table}

\subsection{Complex Plasma Physics}

The second dataset we consider is the plasma dataset \cite{kutz2024shallow}. The data is a time-series of 2,000 time-steps where each time-step is a 14-dimensional grid of 257 $\times$ 256 points. We perform rSVD on each dimension, keeping the most significant 20 modes, in order to reduce the state-space from 921,088 down to a 280 dimensional Reduced Order Model (ROM) \cite{tomasetto2025reducedordermodelingshallow}. The dataset's total size is 785MB.

\begin{figure}[H]
\centering
\includegraphics[width=0.95\textwidth]{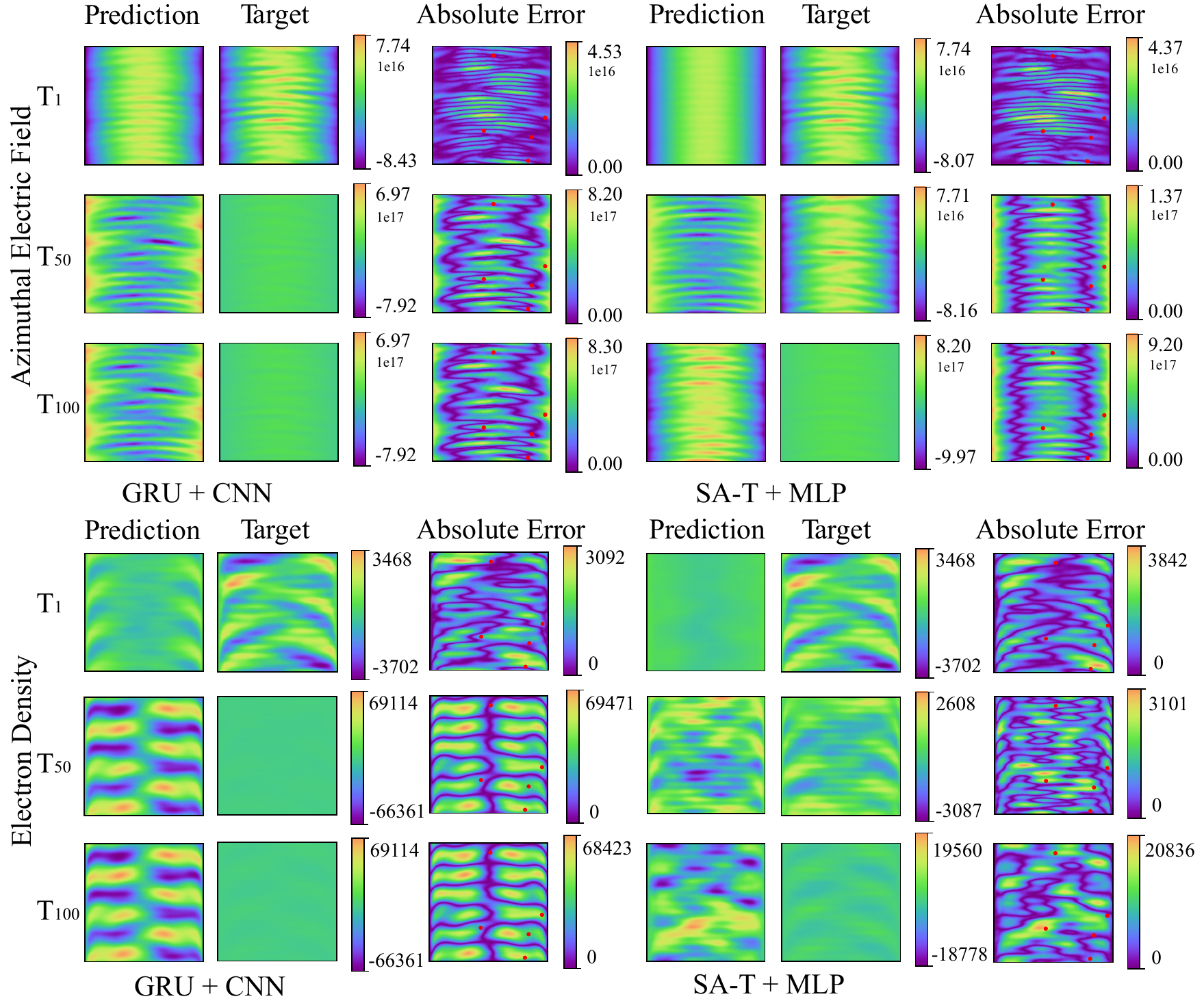}
\caption{Absolute error between the predicted and true values for the best performing SHRED and T-SHRED models on the Plasma dataset. Shown is the azimuthal electric field (top half) and electron density (bottom half). Each row shows a different timestep in the testing dataset. The first three columns show the prediction, target, and absolute error of SHRED with a GRU encoder and a CNN decoder.  The last three columns show the prediction, target, and absolute error of T-SHRED with a SINDy-Attention encoder and an MLP decoder. Sensor locations are in red. \ayy{We use one color scale per prediction-target pair to improve the visual comparison between results.}}\label{fig:plasma_forecast}
\end{figure}

\ay{The lowest average RMSE on the test split of $2.19 \cdot 10^{-2}$ ($\pm\ 1.06 \cdot 10^{-4}$) was achieved from SHRED with a GRU encoder and a CNN decoder with a model size of 0.07 MB. The best performing T-SHRED model was a SINDy-Attention transformer encoder and an MLP decoder with an average RMSE on the test split of $2.29 \cdot 10^{-2}$  ($\pm\ 2.07 \cdot 10^{-5}$) and an average model size of 0.11 MB. A comparison of $100$ steps of forecasting between SHRED and T-SHRED is presented in \autoref{fig:plasma_forecast}.}

\subsection{Shallow Water Equations}

The third example comes from The Well \cite{ohana2025welllargescalecollectiondiverse}, which is a comprehensive collection of numerical physics simulations spanning a diverse set of domains. We use a subset of the \texttt{planetswe} datset from The Well to train our SHRED-based models to observe the performance of T-SHRED in the high-data regime.

The \texttt{planetswe} dataset was generated from simulating the rotating forced hyperviscous spherical shallow water equations:

\begin{align}
    \frac{\partial \textbf{\textit{u}}}{\partial t} = -\textbf{\textit{u}} \cdot \nabla \textbf{\textit{u}} - g \nabla h - \nu \nabla ^4 \textbf{\textit{u}} - 2 \Omega \times \textbf{\textit{u}}, \\
    \frac{\partial h}{\partial t} = -H \nabla \cdot \textbf{\textit{u}} - \nabla \cdot (h \textbf{\textit{u}}) - \nu \nabla ^4 h + F, 
\end{align}
where $\textbf{\textit{u}}$ is the vector-valued velocity field, $h$ is the surface height, $\nu=1.76 \cdot 10^{-10}$ was used for simulation stability, and $F$ is a time-dependent forcing term. We only used the first 10 of the 120 total tracks in the dataset, resulting in a total dataset size of 15.5 GB. Each track contains $1,008$ time-steps where each time-step is a 3-dimensional grid of $256 \times 512$ points.

\ay{The lowest average RMSE on the test split of $5.85 \cdot 10^{-2}$ ($\pm\ 1.32 \cdot 10^{-3}$) was achieved from SHRED with a GRU encoder and an MLP decoder with a model size of 31.50 MB. The best performing T-SHRED model was a SINDy-Attention transformer encoder and an MLP decoder with an average RMSE on the test split of $7.26 \cdot 10^{-2}$ ($\pm\ 2.00 \cdot 10^{-4}$) and a model size of 31.62 MB. A comparison of 100 steps of forecasting between SHRED and T-SHRED is presented in \autoref{fig:planetswe_forecast}. The SINDy-Attention coefficients for one of the seeds resulted in the following set of ODEs in the latent space for the SINDy-Attention layer $L$ and heads $H_j$:}

\begin{align*}
L
\begin{cases}
H_0
\begin{cases}
\dot z_0 = 0.185i \cdot z_0 + 0.192i \cdot z_1 + 0.374i \cdot z_2 -0.147i \cdot z_3\\
\dot z_1 = 0.192i \cdot z_0 + 0.149i \cdot z_1 + 0.544i \cdot z_2 -0.419i \cdot z_3 + 0.669i \cdot z_4 \\
\dot z_2 = 0.374i \cdot z_0 + 0.544i \cdot z_1 -0.386i \cdot z_2 -0.211i \cdot z_3 + 0.617i \cdot z_4 \\
\dot z_3 = -0.147i \cdot z_0 -0.419i \cdot z_1 -0.211i \cdot z_2 + 0.641i \cdot z_4 \\
\dot z_4 = 0.669i \cdot z_1 + 0.617i \cdot z_2 + 0.641i \cdot z_3 -0.115i \cdot z_4
\end{cases}\\
\\
H_1
\begin{cases}
\dot z_0 = -0.343i \cdot z_0 + 0.659i \cdot z_2 + 0.453i \cdot z_3\\
\dot z_1 = -0.388i \cdot z_1 -0.250i \cdot z_3 + 1.433i \cdot z_4 \\
\dot z_2 = 0.659i \cdot z_0  + 0.164i \cdot z_2 + 0.218i \cdot z_3 -0.399i \cdot z_4 \\
\dot z_3 = 0.453i \cdot z_0 -0.250i \cdot z_1 + 0.218i \cdot z_2 + 0.474i \cdot z_3 + 1.454i \cdot z_4 \\
\dot z_4 = 1.433i \cdot z_1 -0.399i \cdot z_2 + 1.454i \cdot z_3 + 0.653i \cdot z_4 \\
\end{cases}\\
\\
H_2
\begin{cases}
\dot z_0 = -0.468i \cdot z_0 + 1.162i \cdot z_1 -0.421i \cdot z_2 + 0.335i \cdot z_3 -0.447i \cdot z_4 \\
\dot z_1 = 1.162i \cdot z_0 + 0.426i \cdot z_1 + 0.122i \cdot z_2  -0.582i \cdot z_4 \\
\dot z_2 = -0.421i \cdot z_0 + 0.122i \cdot z_1 -0.362i \cdot z_2 + 0.735i \cdot z_4 \\
\dot z_3 = 0.335i \cdot z_0 -0.648i \cdot z_3 + 0.331i \cdot z_4 \\
\dot z_4 = -0.447i \cdot z_0 -0.582i \cdot z_1 + 0.735i \cdot z_2 + 0.331i \cdot z_3 -0.839i \cdot z_4 \\
\end{cases}\\
\\
H_3
\begin{cases}
\dot z_0 = -0.227i \cdot z_0 + 0.782i \cdot z_1 -0.259i \cdot z_2\\
\dot z_1 = 0.782i \cdot z_0 -0.694i \cdot z_1 + 0.334i \cdot z_2 + 1.290i \cdot z_4 \\
\dot z_2 = -0.259i \cdot z_0 + 0.334i \cdot z_1  + 0.563i \cdot z_3 \\
\dot z_3 = 0.563i \cdot z_2 + 0.653i \cdot z_3 -0.464i \cdot z_4 \\
\dot z_4 = 1.290i \cdot z_1 -0.464i \cdot z_3 -0.737i \cdot z_4 \\
\end{cases}
\end{cases}
\end{align*}

\subsection{Results}
\label{sec:results}

\ay{From the results on next-step state prediction, we observe that transformers fall behind RNNs. For all datasets, the lowest mean test split RMSE was achieved by the GRU-based SHRED models. These results align with previous research that suggests transformer-based architectures might not be the best for modeling temporal predictions \cite{ekambaram2024tiny,chen2022transformers,wang2025learning}. We also point out that while the \texttt{planetswe} dataset has 15.5GB of data, the encoder only sees less than 0.004\% of the full data due to the sparse sensor measurements ($5 \times 3$ points for every $256 \times 512 \times 3$ input). In order for the transformer encoder to see 15.5GB, we would need to scale up our input dataset to be on the order of 400TB. There is empirical evidence from other transformer foundation models that going to such large scales would greatly improve T-SHRED's performance, however, we do not do so here due to computational and time constraints \cite{wiesner2025towards, chen2024towards}. Furthermore, passing 15.5GB through the transformer is a significantly smaller amount of data than is typically seen in modern foundation models which often pass over 1TB of data in their transformer blocks.}

% 
% 15.5 -> 0.000038
% 406323.2 -> 15.5

\ay{Comparing the T-SHRED models with one another in \autoref{fig:scatter}, we see that the SINDy-Attention models outperform the other T-SHRED models across the \texttt{planetswe} and Plasma datasets over mean test split RMSE, but not on the Sea-Surface Temperature dataset. However, there is significant variability across seeds and none of the datasets demonstrate a clear winner for T-SHRED model architectures. Furthermore, it is not clear if there is any meaningful difference between choosing a CNN or an MLP as a decoder across all models.}

\begin{figure}[H]
\centering
\includegraphics[width=\textwidth]{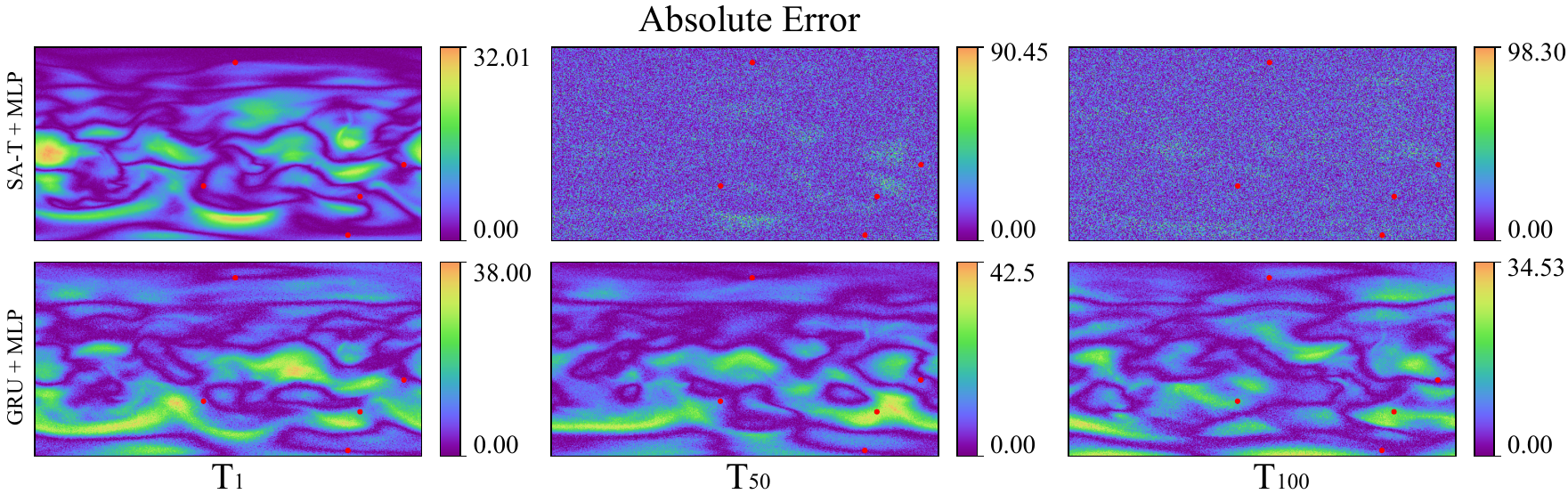}
\caption{Absolute error between the predicted and true heigh for the best performing SHRED and T-SHRED models on the \texttt{planetswe} dataset. The top row shows the absolute error of T-SHRED with a SINDy-Attention encoder and an MLP decoder. The bottom row shows the absolute error of SHRED with a GRU encoder and an MLP decoder. The columns represent the timesteps in the testing set for which the models were evaluated. Sensor locations are in red.}\label{fig:planetswe_forecast}
\end{figure}

From the ODEs in the latent space on the \texttt{planetswe} dataset, we observe that each head of each layer of the SINDy-Attention encoder learns a different ODE in the latent space. This tells us that as the data passes through the encoder, the dynamics of the latent space change. This informs us that the encoder layers manipulate the data into a form that is most accessible for the decoder to generate the full state space as an output. \ay{One of the clear benefits of the SINDy-Attention mechanism is from one-shot rollouts in the latent space. As observed in Figures \ref{fig:sst_forecast}, \ref{fig:plasma_forecast}, and \ref{fig:planetswe_forecast} The long-term forecasting does not fall too far behind in terms of absolute error in forecasting 1, 50, and 100 steps.}

A significant benefit of SHRED models that has been overlooked in the literature is the small model size. We highlight that all of the models were significantly less than 1GB in final size. Despite this relatively small size compared to other deep models in the literature, SHRED is effective at producing full state-space output by learning the underlying dynamics of the input dataset.

\section{Conclusion}
\label{sec:conclusion}

In conclusion, this work suggests a promising direction for combining advanced deep learning technologies from an applied physical and mathematical framework, specifically with the interpretability and regularization of symbolic expressions. Through our carefully designed experiments, 
we demonstrate T-SHRED as 
%the most effective and 
a general architecture for state-space prediction from a sparse set of sensor measurements for larger datasets. 
T-SHRED provides an interpretable model in the latent space by combining SINDy-Attention transformers for symbolic regularization. 
This work also highlights how symbolic regression might be explicitly embedded in other deep learning architectures as a regularizer in order to improve interpretability and accuracy.

% We perform an extensive comparison of SHRED models by combining a zoo of deep learning blocks and find that T-SHRED outperforms all previously used SHRED architectures. 
We study a variety of classic deep learning model blocks, including LSTMs, GRUs, MLPs, and CNNs. The nexts-step forecasting results highlight how traditional SHRED architectures with RNNs outperform T-SHRED models. This highlights a fruitful direction of future work, primarily in identifying key architecture changes to improve the accuracy of T-SHRED models over previous models.

\ay{Importantly, the current work is not focused on extracting the most accurate neural network model for next-step prediction.  Indeed, many sequence models can do this quite well.  And given enough training data and a large enough network, almost any of them can perform at a state-of-the-art level.  We are interested instead in developing T-SHRED as a tool for investigation, diagnostics and characterization.  In the data presented, which features complex, multiscale spatio-temporal dynamics, there are no ground truth models aside from large scale simulations or data collection.  T-SHRED, however, can clearly find simple dynamical models, which have closed form solutions, in its transformer latent space that are capable of representing the multiscale physics observed.  \ayy{While further analysis is needed to fully interpret the learned ODE,} from a scientific perspective this gives a valuable tool to the practitioner for helping probe the system and learn the characteristics of the physics.  There is clearly more work that needs to be done in understanding what the transformer is learning.  But it at least is giving information back to a human in terms of expressions that are typically used to model physics-based systems, i.e. governing equations.  Our aim is to provide a complimentary analysis tool that can aid in the rapid growth of transformer models in science and engineering.  As such, we have demonstrated} a novel advancement in the interpretability of the transformer architecture. In particular, we reformulate the attention heads of the transformer with SINDy-Attention to directly increase interpretability of the model without the cost of performance. The model then learns a set of coupled ODEs that evolve over time in the latent space, providing insight into the dynamics of the dataset that are learned for the task of forecasting full state predictions. This technique presents a paradigm shift in how deep learning can be performed in physics-based settings.

As deep learning advances as a field, interpretability has become an increasing concern. Models with billions of parameters are powerful, yet industry and governments hesitate to use them in practice due to a lack of transparency. It's not clear most of the time what exactly a model is learning, especially when interpretability is treated as a second-class citizen. This work demonstrates that it is possible to progress dynamical systems research in deep learning while also putting interpretability in the foreground. T-SHRED with SINDy-Attention makes no sacrifices in model expressivity, function approximation capacity, or compute cost compared to other T-SHRED architectures. It remains effective, general, and introduces interpretability in a straight-forward manner. The open-source code provided allows for reproducible and broad usage across the sciences. 

\section{Acknowledgements}
\label{sec:ack}

This work was supported in part by the US National Science Foundation (NSF) AI Institute for Dynamical Systems (dynamicsai.org), grant 2112085.  JNK further acknowledges support from the Air Force Office of Scientific Research  (FA9550-24-1-0141). AY is supported by the NSF Graduate Research Fellowship Program under Grant No. DGE-2140004.  Any opinions, findings, and conclusions or recommendations expressed in this material are those of the author(s) and do not necessarily reflect the views of the sponsors.


\bibliography{refs}
\bibliographystyle{plain}

%\newpage 
%\appendix
%\input{arxiv/appendix}

\end{document}